\documentclass[10pt,twocolumn,letterpaper]{article}

\usepackage{iccv}
\usepackage{times}
\usepackage{epsfig}
\usepackage{graphicx}
\usepackage{amsmath}
\usepackage{amssymb}
\usepackage[normalem]{ulem}

\usepackage{multirow}
\usepackage{diagbox}
\usepackage{booktabs}
\usepackage{xcolor}
\definecolor{mygreen}{HTML}{2CB600}
\usepackage{subfig}
\usepackage{xspace}
\usepackage{appendix}
\usepackage[accsupp]{axessibility}  
\usepackage[breaklinks=true,bookmarks=false]{hyperref}
\usepackage[capitalize]{cleveref}

\usepackage[section]{placeins}

\iccvfinalcopy 

\def\iccvPaperID{12510} 

\ificcvfinal\fi

\usepackage[section]{placeins}

\newcommand{\resolved}[1]{}

\newcommand{\draftcomment}[3]{{\textcolor{#3}{[#1]#2}}}
\renewcommand{\draftcomment}[3]{}  

\begin{document}

\newcommand{\datasetname}[0]{\textsc{Whoops!}\xspace}
\newcommand{\website}[0]{\url{whoops-benchmark.github.io/}}
\newcommand{\wizardhat}[0]{\includegraphics[width=.06\textwidth]{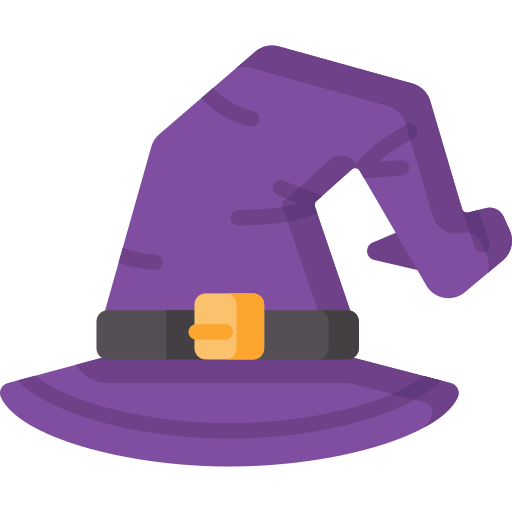}}


\title{Breaking Common Sense: WHOOPS! \\ A Vision-and-Language Benchmark of Synthetic and Compositional Images}

\author{
 \textbf{Nitzan Bitton-Guetta}$^{\ddagger}$\footnotemark[1] \;\; 
\textbf{Yonatan Bitton}$^{\dagger}$\thanks{Equal contribution.} \;\;
\textbf{Jack Hessel}$^{\mp}$ \;\;
\\
\textbf{Ludwig Schmidt}$^{\pm}$ \;\;
 \textbf{Yuval Elovici}$^{\ddagger}$ \;\; 
 \textbf{Gabriel Stanovsky}$^{\dagger,\mp}$ \;\;
 \textbf{Roy Schwartz}$^{\dagger}$\\
 $^{\dagger}$The Hebrew University of Jerusalem \quad
 $^{\ddagger}$Ben Gurion University of the Negev \\
 $^{\mp}$Allen Institute for Artificial Intelligence  \quad
 $^{\pm}$University of Washington \\
\{yonatan.bitton,gabriel.stanovsky,roy.schwartz1\}@mail.huji.ac.il;\\ 
 {nitzangu,elovici@bgu.ac.il}; 
 jackh@allenai.org;
 schmidt@cs.washington.edu\\
}

\maketitle
\ificcvfinal\thispagestyle{empty}\fi

\begin{abstract}
Weird, unusual, and uncanny images pique the curiosity of observers because they challenge commonsense. For example, an image released during the 2022 world cup depicts the famous soccer stars Lionel Messi and Cristiano Ronaldo playing chess, which playfully violates our expectation that their competition should occur on the football field.$^1$ 
Humans can easily recognize and interpret these unconventional images, but can AI models do the same?
We introduce \datasetname{}, a new dataset and benchmark for visual commonsense. The dataset is comprised of purposefully commonsense-defying images created by designers using publicly-available image generation tools like Midjourney.
We consider several tasks posed over the dataset.
In addition to image captioning, cross-modal matching, and visual question answering, we introduce a difficult explanation generation task, where models must identify and explain why a given image is unusual. Our results show that state-of-the-art models such as GPT3 and BLIP2 still lag behind human performance on \datasetname.
We hope our dataset will inspire the development of AI models with stronger visual commonsense reasoning abilities.\setcounter{footnote}{1}
\footnote{Data, models and code are available at the project website: \website{}.}

\end{abstract}

\section{Introduction}
\begin{figure}[!tb]
    \centering
    \includegraphics[width=\columnwidth]{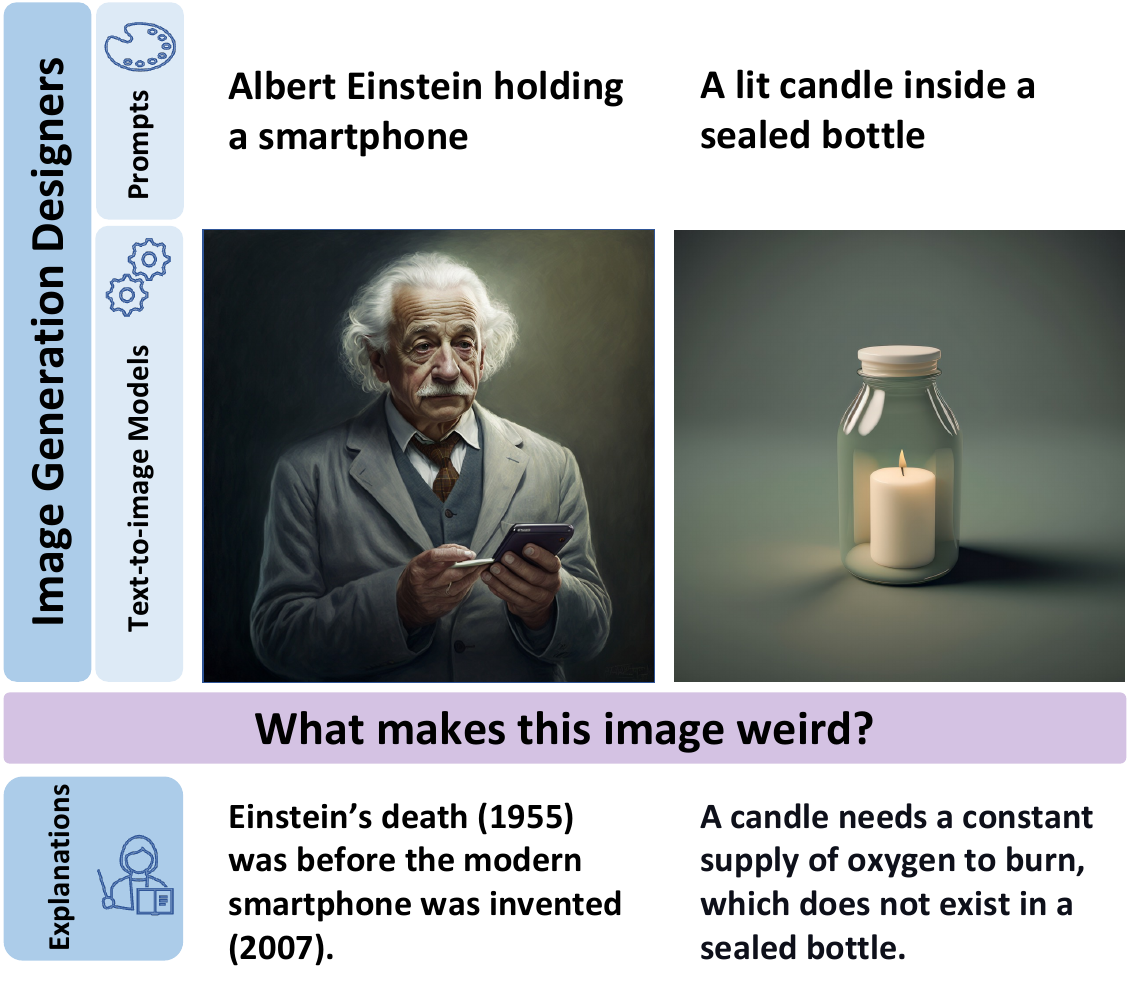}\\
\caption{
We introduce \datasetname{}: a dataset of commonsense-violating images. Designers create interesting, unusual images using prompt-based image-generation tools like Midjourney. We pose several tasks over \datasetname{}, including an explanation generation task. While humans easily identify the weird elements in each image, we show that state-of-the-art AI models struggle.}
    \label{fig:fig1}
\end{figure}

\begin{figure*}[htp]
    \centering
    \includegraphics[width=\textwidth]{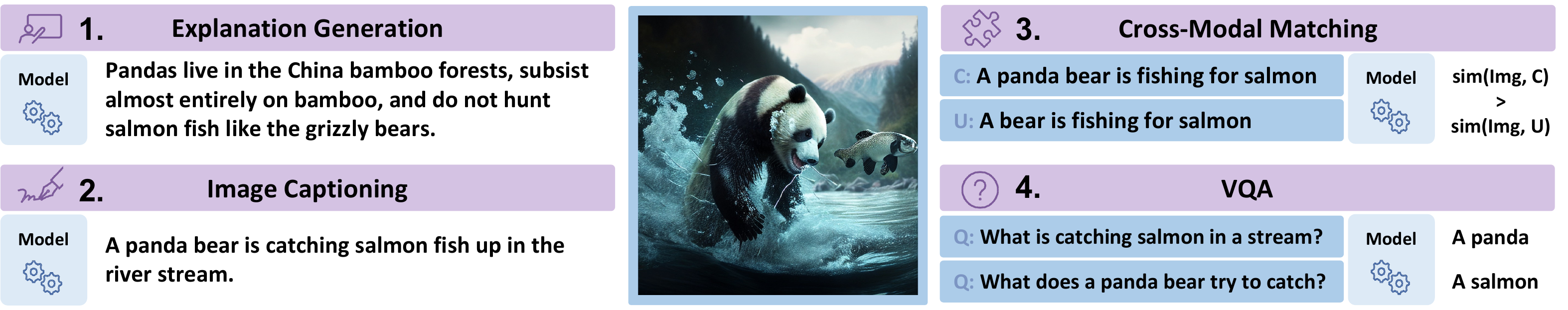}\\
    \caption{The \datasetname benchmark includes four tasks: 1. generating a detailed explanation for what makes the image weird, 2. generating a literal caption, 3. distinguishing between detailed and underspecified captions, and 4. answering questions that test compositional understanding. Inputs to the models are indicated in dark blue.}
    \label{fig:fig_benchmark}
\end{figure*}

Upon viewing an unusual image,
humans can readily recognize odd, unusual, and incongruent factors.
Consider the examples in \cref{fig:fig1}: smartphones did not exist when Einstein was alive (left),
and an oxygen-starved candle would not stay lit for long in a sealed bottle (right). While the images consist of ``normal" constituent objects, \emph{compositions} make them unusual. Although it's relatively easy for humans to identify/explain why an image is unusual, the multi-step reasoning is sophisticated. Connecting visual cues to knowledge about the world goes beyond object recognition, and requires commonsense derived from everyday experiences, physical/social knowledge, and cultural norms~\cite{sap2019socialiqa,zellers2019recognition,hessel2022androids,talmor2022commonsenseqa}.  

In this work, we introduce \datasetname,\footnote{\textit{\textbf{W}eird and \textbf{H}eterogene\textbf{O}us \textbf{O}bjects, \textbf{P}henomena, and \textbf{S}ituations.}} a dataset of 500 synthetic images and 10,874 annotations designed to challenge AI models' ability to reason about commonsense and compositionality. To construct \datasetname, we collaborate with designers who use text-to-image models such as Midjourney, DALL-E \cite{ramesh2021zero} and Stable-Diffusion \cite{rombach2022high} to generate images that would be challenging (or even impossible) to collect otherwise. First, prompts that contain two plausibly co-occurring elements are constructed, and then, a modification to one of them is made to create an implausible combination that violates commonsense. \cref{fig:fig1} (left), for example, was created by our designers thinking of a plausible scene of Albert Einstein holding a notebook, and then replacing the notebook with a smartphone, which did not exist at the time. We annotate our images with textual information, including both descriptive captions and explanations for what makes each image weird.

Next, we pose four visual commonsense reasoning tasks over the \datasetname{} corpus: (1) explanation generation, where models provide detailed explanations of what makes an image weird; (2) image captioning, where models summarize the content of the images; (3) cross-modal matching, where models should score a detailed caption higher than a correct but underspecified one, and (4) visual question answering, where models answer questions that test their comprehension of the weird images (Fig.~\ref{fig:fig_benchmark}). Our evaluation covers both zero-shot and supervised experimental settings.

Experiments on \datasetname{} show that state-of-the-art vision-and-language models (e.g.,  OFA~\cite{wang2022ofa}, BLIP~\cite{li2022blip}, CoCa~\cite{yu2022coca}) lag behind human performance for all tasks. For instance, a human evaluation reveals that a fine-tuned version of BLIP2-XXL~\cite{li2023blip} achieves a performance of 27\% acceptability, and a ``pipeline'' approach of feeding a predicted image description to the latest version of GPT3 (davinci-003) \cite{brown2020language} reaches 33\%. However, both these models fail to generate explanations as well as humans, who achieve 95\% on the same task. 

To support fully automated evaluations, we present a model-based metric for the explanation-of-violation task. This involves a GPT4 model on ground-truth explanation and predicted explanation. Achieving an accuracy of over 81\%, this metric aligns well with human ratings. We make human annotated data and the complete automatic evaluation code publicly accessible. Researchers can evaluate their models and submit the results to the leaderboard on the project website.

Finally, we show that the difficulty \datasetname{} goes beyond recognition; even providing a ground-truth oracle image description instead of the predicted caption in the "pipelined" setting, models still struggle to effectively explain the incongruity of the scene, with an accuracy rate of only 68\%. Overall, our results show that \datasetname{} is a challenging benchmark, even for state-of-the-art vision-and-language models. This result highlights the need for continued development in commonsense reasoning, compositionality, and explanation generation. We release our models, code, and data.

\section{Collecting \textit{Weird} Images}
\label{sec:image_generation}

\datasetname is designed to challenge vision-and-language models with images that require commonsense reasoning and understanding beyond simple object co-occurrence. 
The term ``weird'' is ultimately subjective, ambiguous, and culture-dependent. Because our goal is to create a benchmark, we aim to generate images that are unusual for a diverse set of reasons, including temporal, biological, cultural, physical and others.  
We start by describing how we generate the images, and then present an analysis of the different reasons for the images, which shows that our dataset is indeed diverse in this respect.

\begin{figure*}
    \includegraphics[width=\textwidth]{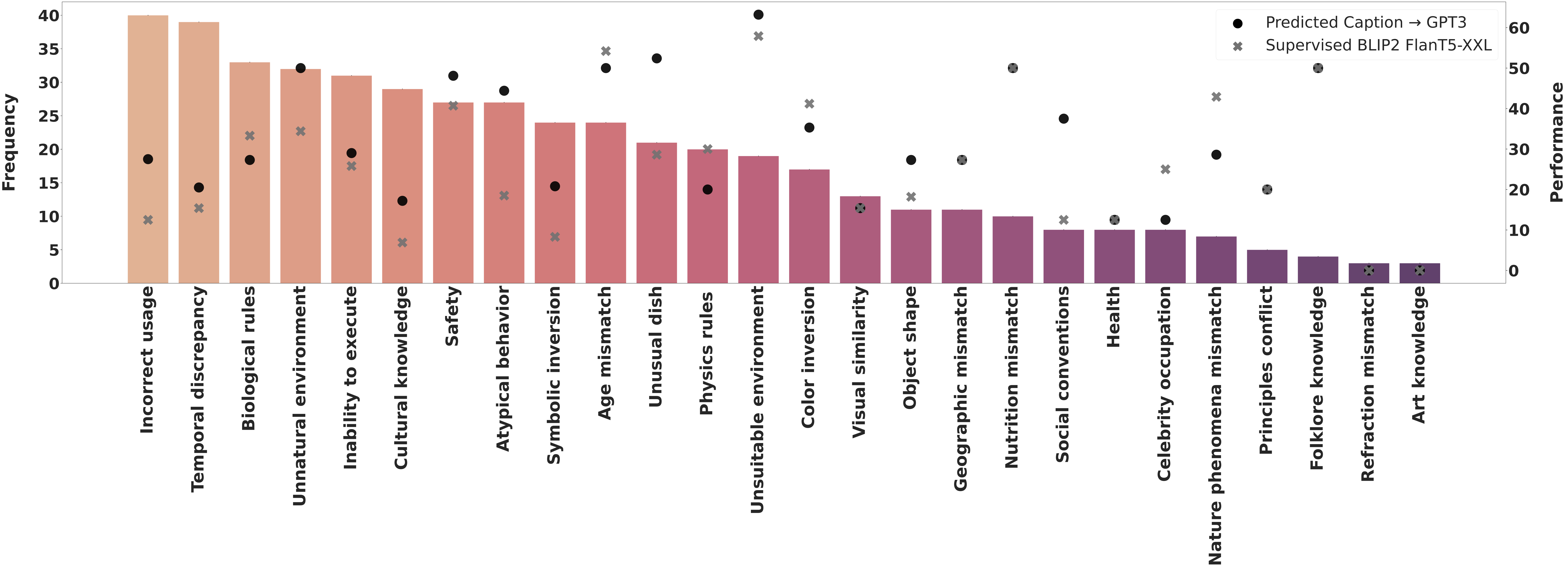}
    \caption{A histogram of the annotated commonsense reasons for \datasetname images to be weird. The reasons include a wide range of deviations from expected social norms and everyday knowledge. We also present the explanation generation performance of the two top models in our experiments (\cref{sec:experiments}). Left axis is the frequency for each commonsense category, and right is the performance of both models. 
    }
    \label{fig:clusters}
\end{figure*}


\begin{figure*}[!ht]
    \centering
    \includegraphics[width=\textwidth]{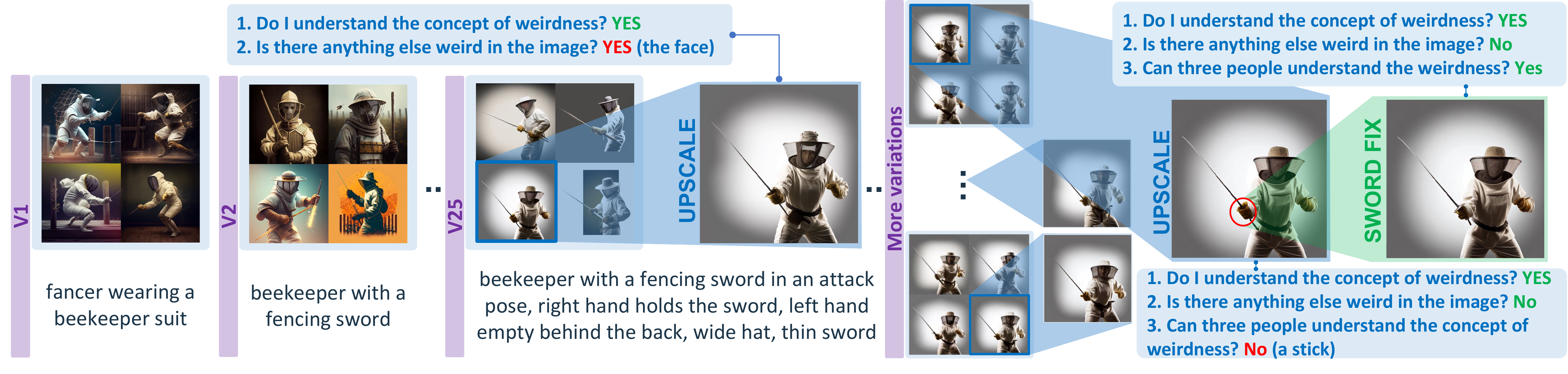}
    \caption{\datasetname image generation example: This image, produced through more than 25 iterations, demonstrates the process from initial prompt to finalized 'weird' image produced by the text-to-image model. In each iteration the designer verifies (1) their understanding of the 'weirdness' concept, (2) absence of any additional weird elements in the image, and (3) the 'weirdness' clarify to three independent individuals. Only images meeting these criteria are included in the dataset.}
    \label{fig:fig_create_img}
\end{figure*}

\subsection{Human Generated Synthetic Images via Text-to-Image Models}
We recruit a group of 30 image designers who using  Midjourney, DALL-E \cite{ramesh2021zero}, or Stable-Diffusion \cite{rombach2022high} as text-to-image models. They are requested to generate weird images by first coming up with ``weird'' prompts, and editing them until a desired image is generated. 

These prompts should adhere to the following guideline: first generate a prompt of an image that depicts two elements that are \textit{likely} to co-occur, and then replace one of them with a different element to create a new prompt that describes an image that is \textit{unlikely} to exist in reality. For instance, taking a prompt of Albert Einstein holding a notebook and replacing the notebook with a smartphone, resulting in a prompt for an unlikely image, as smartphones did not exist during Einstein's time.
Each image is required to be synthetic, rather than edited from existing images. This results in a total of 500 weird images. See \cref{sec:image_generation_guidelines} for full guidelines and examples.

The generation of each image, as depicted in \cref{fig:fig_create_img}, goes beyond a simple use of a text-to-image model. It is a meticulous process, managed by expert designers, involving around 25 iterations per image. The primary objectives are to ensure the image clearly portrays its 'weirdness' concept and to eliminate any extraneous elements that could be misunderstood as the main source of 'weirdness.'

For every image, the designers also provide a concise one-sentence explanation that encapsulates the unique reason for its 'weirdness.' The authors review and refine these explanations for factual accuracy and specificity, incorporating relevant details such as names, dates, and other pertinent information.

This process aims to create images with unmistakable 'weirdness,' understandable to a wide audience. To verify this, each image is presented to a small control group before its inclusion in the dataset. Images that fail to pass this test are returned to the designers for refinement or to explore alternative concepts. To mitigate the concern that weirdness is limited to a specific culture or region, the designers who created the images and the annotators who labeled the data (\cref{sec:benchmarking}) are from different countries and continents.


\subsection{Commonsense Categorization of \textit{\textbf{Weird}} Images}\label{sec:image_analysis}

\cref{fig:clusters} provides a histogram of the different types of commonsense reasoning that underlie the weirdness of images in \datasetname. To create it, we manually annotate each image with the main reason that contributes to its overall sense of ``weirdness''. Our annotation includes 26 different categories. The reasons cover a broad range of domains, including but not limited to temporal discrepancy (\cref{fig:fig1} left), physical rules (\cref{fig:fig1} right), nutrition mismatch (\cref{fig:fig_benchmark}), unsuitable environment (\cref{fig:figQA}), atypical activity (\cref{fig:explanation_baselines}), symbolic inversion, folklore knowledge and more. This analysis shows the diversity and complexity of the reasoning skills that vision-and-language models must possess in order to perform well on our benchmark. 
Further elaboration on commonsense categories is available in \cref{sec:commonsense_categories}, including examples.

\section{Related Work}
The field of commonsense reasoning has recently gained significant attention, with various tasks proposed both in natural language processing (NLP) \cite{saha2021explagraphs, zellers2018swag, zellers2019hellaswag, sap2019atomic, bisk2020piqa, forbes2019neural} and computer vision \cite{vedantam2015learning, bitton2022vasr}. In the field of vision-and-language, models are being developed to solve complex visual reasoning tasks. These include visual understanding tasks, such as VCR \cite{zellers2019recognition}, as well as tasks that evaluate commonsense reasoning in association and analogy tasks, like WinoGAViL \cite{bitton2022winogavil} and VASR \cite{bitton2022vasr}. Other tasks evaluate compositionality (e.g.,  Winoground;  \cite{thrush2022winoground}), visual abductive reasoning (e.g., Sherlock; \cite{hessel2022abduction}) and comprehension and explanation of multi-modal humor \cite{hessel2022androids}.
Recent progress in large language models is making way for models that can solve these tasks using instructions like BLIP2 \cite{li2023blip} and in-context learning, or zero-shot learning, like Flamingo \cite{alayrac2022flamingo} and MLLM \cite{https://doi.org/10.48550/arxiv.2302.14045}. These recent advances pave the way for our work, which provides a challenging resource for commonsense and compositionality.

\datasetname, is distinct from prior work that focuses on reasoning with pre-existing images. Instead, it contains synthetic images that are specifically designed to challenge AI models' abilities to reason about commonsense and compositionality, with an emphasis on images that violate expectations. Our approach uses image generation models to create unique and complex images that would be difficult or impossible to obtain otherwise, providing an opportunity to evaluate critical aspects of visual reasoning, including compositionality and commonsense reasoning.

\section{Using Weird Images to Create V\&L Tasks}
\label{sec:benchmarking}
We pose four vision-and-language tasks over the \datasetname{} images to form a benchmark. 
We first consider a novel task---\textit{explanation-of-violation generation}---which evaluates a model's ability to identify the commonsense rule that an image violates and reason about the relationships between different elements in the image. 
We then consider three other well-established tasks posed over the corpus: image captioning, cross-modal matching, and VQA. \cref{fig:fig_benchmark} shows example annotations for the four tasks over a single image. Previous works have shown that these tasks can be prone to relying on language priors \cite{jabri2016revisiting, zhang2016yin, goyal2017making, agarwal2020towards, bitton2021automatic, bitton2021data, dancette2021beyond}. However, the images in \datasetname were purposefully designed to include uncommon combinations, which may make it more challenging for models to exploit language priors. We also report the (relatively low) performance of a text-only baseline in \cref{sec:results}. The end result is a challenging test set for evaluating the performance of vision-and-language models on complex reasoning tasks. 

To create task instances, we crowdsource annotations for each of the 500 images in our dataset, including captions, various explanations. We pay each annotator 12--15\$ per hour for providing a caption and a explanation; annotation details/instructions are available in the \cref{app:human_annotation}. We also use auto-generation techniques to create VQA data based on these captions \cite{changpinyo2022all,honovich2021q}, and then validate it using human verification. We describe each task and its evaluation below.

\subsection{Explanation-of-violation Generation}
\label{sec:explanation_generation}
The task of the explanation generation is to provide a single-sentence detailed explanation of what makes an image weird. The goal is to test a model's ability to identify the commonsense rule that the image violates and reason about the relationships between different elements in an image. For instance, in \cref{fig:fig_benchmark}, the explanation should provide information such as, \textit{``Pandas usually reside in Chinese bamboo forests, eat almost exclusively bamboo, and do not hunt salmon fish like grizzly bears do"}. We break the task down into two components: \textit{identifying} whether an image is weird and \textit{explaining} what makes it weird. 

\paragraph{Identifying weird images.}
We select a subset of 100 weird images and use a similar protocol to the one described in \cref{sec:image_generation} to collect the corresponding ``normal'' images for them (e.g., for the image of Einstein holding a smartphone, generate an image of him holding a notebook). This task is evaluated using binary accuracy over this paired set, where random chance is 50\%. To assess human performance on this task, we ask three human annotators to classify each image as either ``weird'' or ``normal''. We determine the final classification through a majority vote. Human performance is 92\%, and 3/3 agreement is achieved in 70\% of the cases. These results suggest that, while ``weirdness" is subjective, on average, humans readily agree on what is weird and what is not in the context of \datasetname{}.

\paragraph{Explanation-of-violation.}
We ask annotators to
provide a detailed single-sentence explanation of what makes the image strange and include the reason why two elements are unlikely to co-exist in the scene. We collect five explanation per image, a total of 2,500. The metric to evaluate model predictions on this task is human judgment. We compare model generations to references using three crowdworker judgments: full details and examples in \cref{app:human_annotation}.  

\subsection{Established Tasks}\label{sec:established}
\paragraph{Image captioning.}
This task requires generating a single-sentence description of an image that includes both elements whose combination makes the image weird. Unlike the explanation task, the captioning task does not demand any reasoning about incongruities. i.e., for the example in \cref{fig:fig_benchmark}, it suffices to just generate \textit{A panda bear fishing for a salmon in the river}. This identification task, however, could be helpful for the explanation task presented in \cref{sec:experiments}. 
We crowdsource five textual descriptions per image, for a total of 2,500 captions;
evaluation is using the standard automatic captioning metrics CIDEr \cite{vedantam2015cider} and BLEU-4 \cite{papineni2002bleu} compared to crowd-authored references. 

\paragraph{Cross-modal matching.}
In this task, a model is given an image and a set of captions, all of which accurately describe the scene, but some of which leave out important details. The evaluation setup challenges models to rank the detailed captions more highly than the underspecified ones. This task tests the model's ability to match the correct caption to the image and overcome its language priors, e.g., a text-only model may rate ``A panda hunting for salmon" less likely than ``A bear hunting for fish".\footnote{We confirm this point with a FlanT5 XL language model \cite{chung2022scaling} by asking it to determine which caption is more likely, and it rates the underspecified one as more likely in 85\% of cases.} 
Performance is measured as the proportion of correct rankings.
We collect 500 underspecified captions per image, a total of 2,500 captions. 

\paragraph{Visual question answering (VQA).}
\begin{figure}[!t]
    \centering
    \includegraphics[width=0.6\columnwidth]{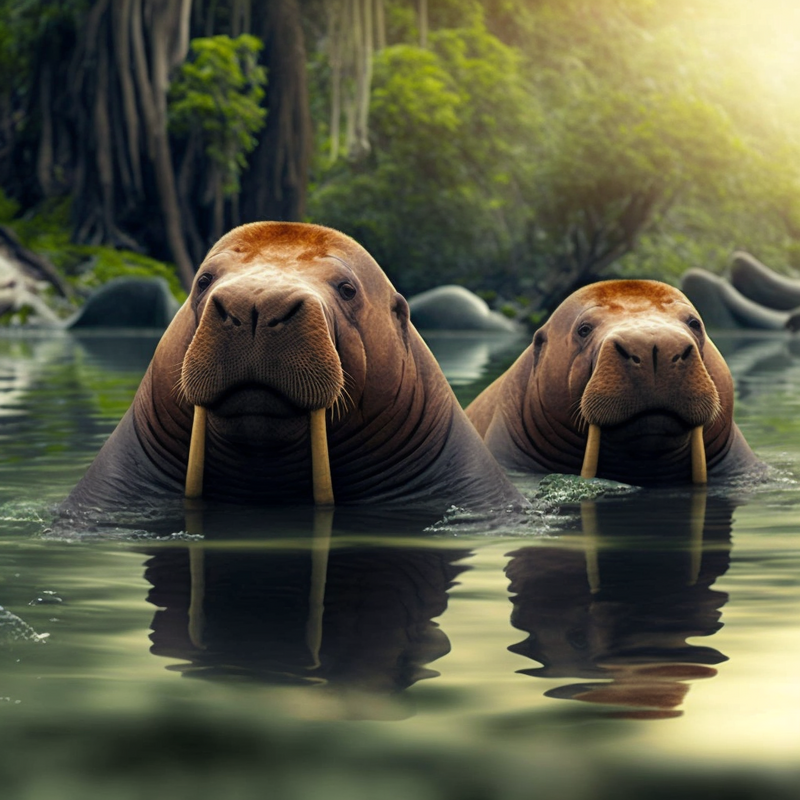}\\
    \caption{We obtain five caption from human annotations, for example: ``Two walruses are swimming in the jungle''. We then automatically generate question-answering pairs: \\
    (1) Where are the two walruses swimming? \textit{in the jungle}\\ 
    (2) How many walruses are swimming in the jungle? \textit{two}\\
    (3) What is swimming in the jungle? \textit{two walruses}}
    \label{fig:figQA}
\end{figure}

To create question-answer pairs for \datasetname{}, we follow the $Q^2$ pipeline for automatic VQA generation \cite{changpinyo2022all,honovich2021q}. This process: 1) derives answers from captions; 2) uses a question generation model to generate a question for each answer; 3) filter the generated questions with the $Q^2$ NLI model. \cref{fig:figQA} presents generated questions and answers for each of the answer candidates in the image caption. We then  \cite{honovich2021q} to ensure that the questions are answerable. We filter out instances solvable by a text-only model performs well so that models must focus on visual-textual interactions. Specifically, we use a language model, FlanT5 XL \cite{chung2022scaling}, to answer the questions and filter out instances where the BEM metric is above 0.1. This filtering removes approximately 30\% of the questions and results in 3,374 VQA samples. In~\ref{sec:results}, we show a text-only finetuned model performs poorly on the resulting set. We evaluate using two metrics: (1) strict exact match; and (2) BERT Matching (BEM) \cite{bulian2022tomayto}, which approximates a reference answer to a candidate answer given a question \cite{devlin2018bert} using a language model score.

We manually verify a sample 300 (image, question, answer) triplet by asking three crowdworkers to classify whether the answer is correct. For a baseline for human verification, we mix in randomly sampled 25\% of the ``negative'' answers. The majority vote is selected as the final answer. Humans reach full agreement in 94\% of the cases, and the majority vote agrees with the automatic VQA label in 97\% of the cases, which provides strong evidence that the generation process generates high-quality QA instances.

\subsection{Toxic Content Filtering}\label{sec:toxic}
Finally, we take two steps to filter toxic and harmful images. First, four of the paper authors manually verify all images and remove those that could be potentially offensive for some groups. Second, we use the Perspectives API\footnote{\url{https://www.perspectiveapi.com/}} to detect and filter out toxic language from our annotated data. We find that the vast majority of our data is non-toxic. Only a very small percentage (0.4\%, 0.1\%, and 0.4\% for captions, explanations, and underspecified captions, respectively) contains toxic language, which we have removed and replaced with new data.

\section{Experiments}
\label{sec:experiments}
\begin{table*}[!htb]
\centering
\begin{tabular}{@{}llcccc@{}}
\toprule
& Task & \multicolumn{1}{p{1.7cm}}{\centering Identify} && \multicolumn{1}{p{1.5cm}}{\centering Explain} \\
\cmidrule(l){3-3} \cmidrule(l){4-6}
& & \multicolumn{1}{p{1.9cm}}{\centering Binary Accuracy $(\uparrow)$} & \multicolumn{1}{p{1.5cm}}{\centering Human Rating $(\uparrow)$} & \multicolumn{1}{p{1.5cm}}{\centering GPT4 Rating $(\uparrow)$} & \multicolumn{1}{p{1.9cm}}{\centering GPT4 Rating Accuracy $(\uparrow)$} \\
\midrule
\multirow{5}{*}{End-to-end} & BLIP2 FlanT5-XXL (Zero-shot) & 50 & \phantom{0}0 & 12 & 88 \\
& BLIP2 FlanT5-XL (Fine-tuned) & 60 & 15 & 18 & 87 \\
& BLIP2 FlanT5-XXL (Fine-tuned) & 73 & 27 & 27 & 81 \\
\cmidrule(l){2-6}
& InstructBLIP & - & - & 31 & - \\
& mPLUG-Owl & - & - & 24 & - \\
& LLaVA & - & - & 31 & - \\
\midrule
\multirow{8}{*}{\begin{tabular}[c]{@{}l@{}}Pipeline\\ (Zero-shot)\end{tabular}} 
& Predicted Caption $\rightarrow$ GPT3 & 59 & 33 & 36 & 87 \\
& Ground-truth Caption $\rightarrow$ GPT3 (Oracle) & 74 & 68 & 70 & 81 \\
\cmidrule(l){2-6}
& Predicted Caption $\rightarrow$ GPT4 & - & - & 36 & - \\
& Predicted Caption $\rightarrow$ Llama-2-7b & - & - & 36 & - \\
& Predicted Caption $\rightarrow$ Llama-2-13b & - & - & 36 & - \\
& Ground-truth Caption $\rightarrow$ GPT4 (Oracle) & - & - & 69 & - \\
& Ground-truth Caption $\rightarrow$ Llama-2-7b (Oracle) & - & - & 71 & - \\
& Ground-truth Caption $\rightarrow$ Llama-2-13b (Oracle) & - & - & 70 & - \\
\midrule
& Humans & 92 & 95 & -& -\\
\bottomrule
\end{tabular}
\caption{
Test results for Explanation-of-violation encompass two main tasks: \emph{identifying} unusual images and \emph{explaining} their anomalies. While humans consistently outperform models across tasks, providing an oracle image description narrows the performance gap. The \emph{explaining} subtask incorporates metrics from human evaluations and GPT4 ratings. These metrics quantify the fraction of correctly classified explanations, either by human judgment or the GPT4 model, with the latter's accuracy detailed in the GPT4 Rating Accuracy column. Models without human evaluation, namely InstructBLIP, GPT4, Llama-2, mPLUG-Owl, LLaVA, were added on August 9, 2023, with the executing of GPT4 auto-evaluation.
}
\label{tab:identification_and_explanation}
\end{table*}

\begin{table*}[!htb]
\centering
\begin{tabular}{@{}llccccc@{}} \toprule
\multicolumn{2}{l}{\multirow{2}{*}{}}                    & \multicolumn{2}{c}{Image Captioning} & \multicolumn{2}{c}{VQA} & Matching    \\ \cmidrule(l){3-4} \cmidrule(l){5-6} \cmidrule(l){7-7} 
\multicolumn{2}{l}{}                                     & B-4 $(\uparrow)$             & CIDEr $(\uparrow)$             & ExactM $(\uparrow)$  & BEM $(\uparrow)$  & Specificity $(\uparrow)$  \\ \midrule
\multirow{5}{*}{Zero-shot}  & CLIP ViT-L/14              & \phantom{0}--                 & \phantom{0}--                & \phantom{0}--                 & \phantom{0}--   & \phantom{0}70          \\
                            & OFA Large                  & \phantom{00}0                 & \phantom{00}0                & \phantom{00}8                 & \phantom{0}38  &             \\
                            & CoCa ViT-L-14 MSCOCO       & \phantom{0}25                & 102              & \phantom{0}--                 & \phantom{0}--   & \phantom{0}72          \\
                            & BLIP Large                 & \phantom{0}13                & \phantom{0}65               & \phantom{00}6                 & \phantom{0}39  & \phantom{0}77          \\
                            & BLIP2 FlanT5-XXL           & \phantom{0}31                & 120              & \phantom{0}15                & \phantom{0}55  & \phantom{0}71          \\ \midrule
\multirow{2}{*}{Fine-tuned} 
                            & BLIP2 FlanT5-XL            & \phantom{0}41                & 174              & \phantom{0}20                & \phantom{0}55  & \phantom{0}81          \\
                            & BLIP2 FlanT5-XXL           & \phantom{0}\textbf{42}                & \textbf{177}              & \phantom{0}\textbf{21}                & \phantom{0}\textbf{57}  & \phantom{0}84          \\ \midrule
                            \multirow{1}{*}{Text only FT} & BLIP2 FlanT5-XXL  & \phantom{00}1                 & \phantom{00}2                & \phantom{00}4                 & \phantom{0}24  & \phantom{0}\textbf{94}          \\

                            \bottomrule
\end{tabular}
\caption{Test results for image captioning, cross-modal matching and visual question answering. A fine-tuned version of BLIP2 FlanT5-XXL generally performs best but there's significant headroom.
}
\label{tab:three_tasks_results}
\end{table*}
\label{sec:methods}
\begin{figure}[!t]
    \includegraphics[width=\columnwidth]{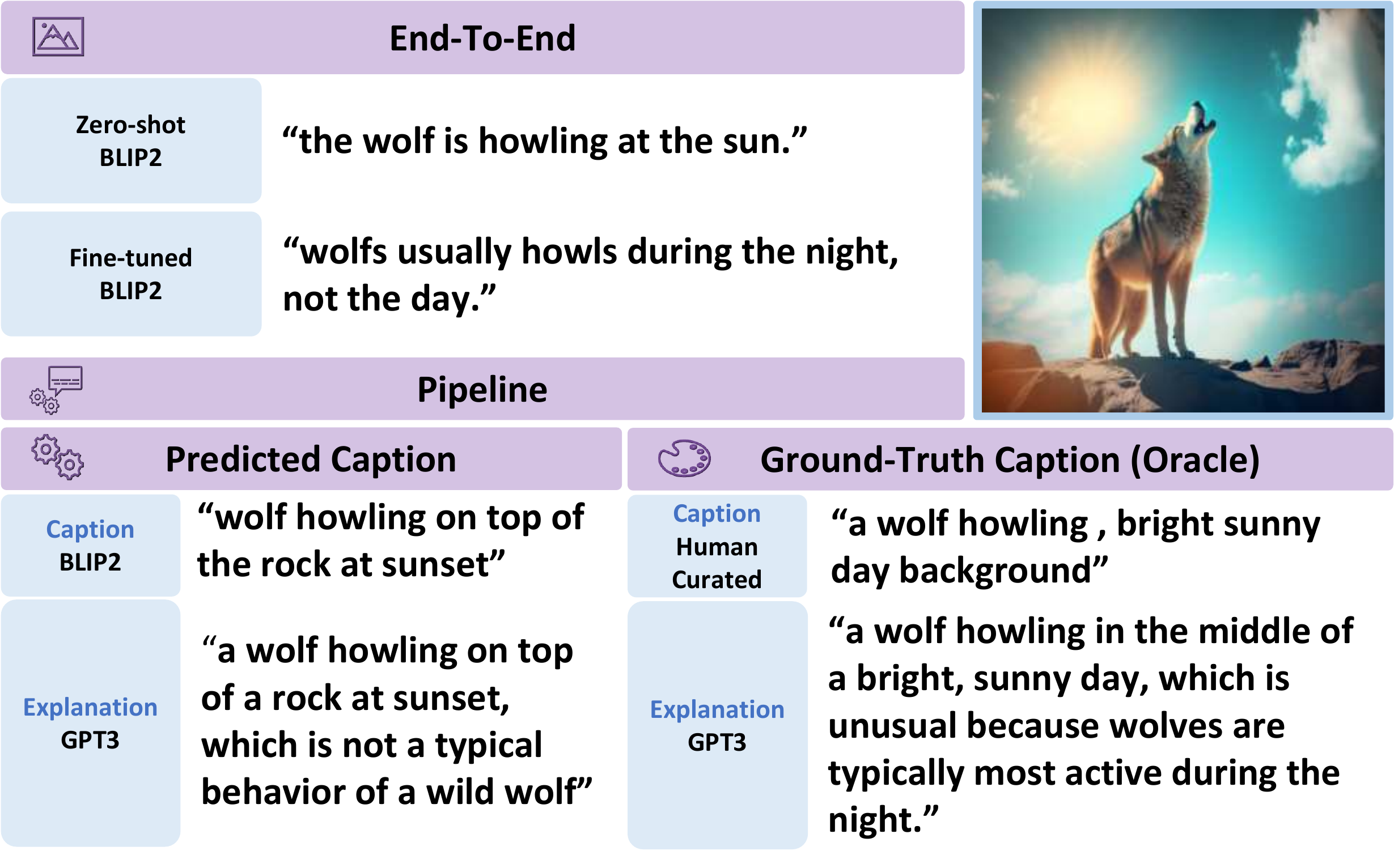}\\
    \caption{We explore two approaches for explanation-of-violation generation. The first approach involved using an end-to-end model that receives an image as input and generates an explanation as output, evaluating both a zero-shot and fine-tuned versions of the BLIP2 model. The second approach was a pipeline that predicted a caption for the image, or used a ground-truth caption (oracle), and then used a language reasoning model (GPT3) to generate an explanation based on the caption.}
    \label{fig:explanation_baselines}
\end{figure}

We evaluate models on our tasks in a fully zero-shot setting and a 5-fold cross-validation supervised configuration. 

\paragraph{For zero-shot evaluations,}
we use the officially published implementations of CLIP ViT L/14~\cite{radford2021learning}, OFA Large~\cite{wang2022ofa}, BLIP Large~\cite{li2022blip}, CoCa ViT-L-14 MSCOCO~\cite{yu2022coca}, and BLIP2 FlanT5-XXL~\cite{li2023blip}. Additional details can be found in \cref{sec:reproduce}. Some models can be used to tackle all tasks (BLIP2), and some only a subset of the tasks: OFA for image captioning and VQA; CoCa for image captioning and cross-modal matching; and CLIP only for cross-modal matching. For CLIP and CoCa, we evaluate all available model versions (four for each model), and for readability report the best performing ones.

\paragraph{For supervised evaluations,} we fine-tune the BLIP2. To report over the same instances as in the zero-shot evaluations, we split the images in \datasetname{} into 5 cross-validation splits. For these 5 splits independently, we train supervised models using 60\% of the data as training, 20\% as validation, and 20\% for test. We fine-tune just the Q-former parameters of BLIP2 using Adam \cite{kingma2014adam}. We train for 15 epochs, and use the validation set for early stopping and to select learning rate between \{1e-5, 5e-5\}. We concatenate training instances in a sequence-to-sequence format for all tasks jointly such that a single supervised model can address all tasks; see \cref{sec:sec_with_supervised_data_details} for details.

\paragraph{We also consider ``pipelined," methods \cite{zeng2022socratic}} for the explanation-of-violation tasks. These methods decouple recognition of objects from reasoning about incongruity. In the ``pipelined" approach, an image caption is passed to a large language model (LLM), which is then tasked with the two explanation-of-violation subtasks. We use GPT3 text-davinci-003 as the LLM \cite{brown2020language} and experiment with two textual descriptions: a predicted image caption by the BLIP2 model and an oracle version, which includes the ground-truth caption collected by annotators and verified by the authors.

\paragraph{As a baseline,} we train a text-only BLIP2 FlanT5-XXL using the same cross-validation/hyperparameter setup as for the full supervised models, except we set all pixels of the image to mean so that image content cannot be used at training or testing time.

\paragraph{Addition of new models for explanation-of-Generation task.}
on August 9, 2023, we expanded our zero-shot evaluations by incorporating new models. For the ``pipelined'' models, we integrated Llama2~\cite{touvron2023llama} and GPT4~\cite{openai2023gpt}. Additionally, for the end-to-end vision-language models, we introduced InstructBLIP~\cite{dai2023instructblip}, LLaVA~\cite{liu2023llava} and mPlug-Owl~\cite{ye2023mplug}.

\subsection{Automatic Evaluation for Explanation-of-violation}
\label{sec:auto_eval_explanations}
In the explanation-of-violation task, we supplement human judgments with an automatic evaluation metric, demonstrating its significant alignment with human assessments. The objective of introducing this automatic evaluation method is to provide a reproducible standard for result reporting on \datasetname{}.

We utilize the human-annotated data as described in \cref{sec:benchmarking} and shown in \cref{fig:explanation_selection}. This data comprises 5,000 pairs of ground-truth explanations, candidate explanations, and an associated label indicating whether humans rated the given explanation as correct or not.


We report two results: (1) The proportion of positive and negative labels. This forms the \emph{model rating} that can be compared with the \emph{human rating}. (2) The accuracy of the automatic evaluation metric as compared to the human judgement. This helps in understanding the alignment between human and machine-based assessments. 


We used a GPT4 model with the prompt: \textit{Evaluate the equivalence of the following explanations for the question "What is unusual in this image?" Answer with True or False: A: $sentence_A$ B: $sentence_B$. True if A and B have the same meaning, False if they do not.}
The parameters $sentence_A$, $sentence_B$ contains the ground truth explanation and the candidate explanation respectively. \footnote{We publish the annotated data and code for automatic evaluation on \datasetname{} in this \href{https://colab.research.google.com/drive/1yphyMKFFrK7TrNVhbVIlzkYfs77mZBxi?usp=sharing}{notebook}.}

\subsection{Results}
\label{sec:results}
\paragraph{Explanation-of-violation.} The results for the two identification and explanation subtasks, are presented in \cref{tab:identification_and_explanation}. For both cases, models significantly lag behind human performance. For example, on identification, the best end-to-end fine-tuned BLIP2 FlanT5-XXL model achieves at best 73\%. For explanation, even the oracle model (which is given access to a ground-truth, human-authored description of the image) only achieves a performance of 68\%, falling substantially short of human performance (95\%). These results indicate that our dataset provides a challenging benchmark for the development of next-generation vision-and-language models. We provide an example of model predictions in \cref{fig:explanation_baselines} and an example of the evaluation task to rate both model predictions and human explanations in \cref{fig:explanation_selection}. 

In the automated explanation of violation task, the maximum deviation between the automatic rating and human rating is 9\% for the Predicted Caption → GPT3 task. All the automated explanation-of-violation models achieve an accuracy higher than 81\%. These results suggest that automatic evaluation yields ratings correlated to human ratings.

\paragraph{Captioning, VQA, + Matching.} 
The results are presented in \cref{tab:three_tasks_results}. The zero-shot results highlight the strengths and weaknesses of each model. OFA achieves the lowest results, particularly in image captioning, where it frequently predicts the pattern ``digital art selected for the \#''. Zero-shot BLIP2 demonstrates a substantial improvement over the other models. But even the supervised models have significant room for improvement, especially in VQA (maximum BEM score is 57\%) and image captioning; In section \ref{sec:analysis}, we conducted an analysis in which humans rate the BLIP2 zero-shot predictions. Despite the relatively high CIDEr score, the model failed to capture important information, resulting in a human rating of 49\%. We also report results for a text-only supervised baseline. The results show that it performs poorly on captioning and VQA.
\footnote{Our question collection process filters out questions that could be answered using text alone in a zero-shot setting (see \cref{sec:established}). The current analysis validates that this filter indeed removed shortcuts, even for supervised models.
In contrast, the text-only model performs well in the matching task with a performance of 94\%: likely, the model learns to prefer more detailed captions,
even without seeing the image. We thus advocate for matching to be used only as a zero-shot evaluation.}


\section{Analysis}
\label{sec:analysis}
\begin{figure}[!ht]
    \centering
    \subfloat[a pair of white ice skates on an ice rink]{\includegraphics[width=0.3\columnwidth,height=2.5cm]{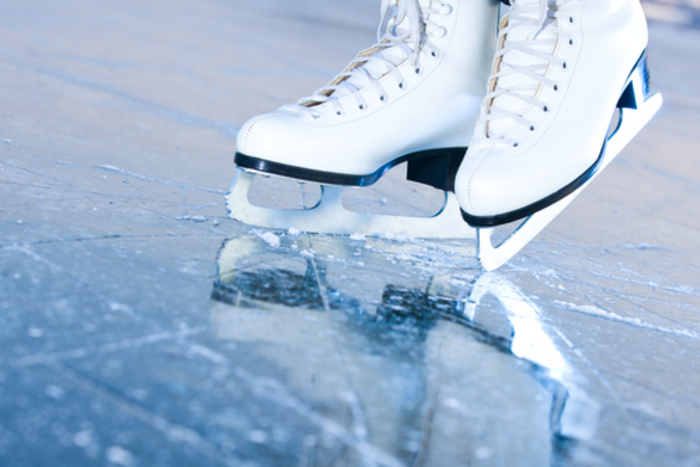}}\hfill
    \subfloat[a close up of a person's skates on an ice rink]{\includegraphics[width=0.3\columnwidth,height=2.5cm]{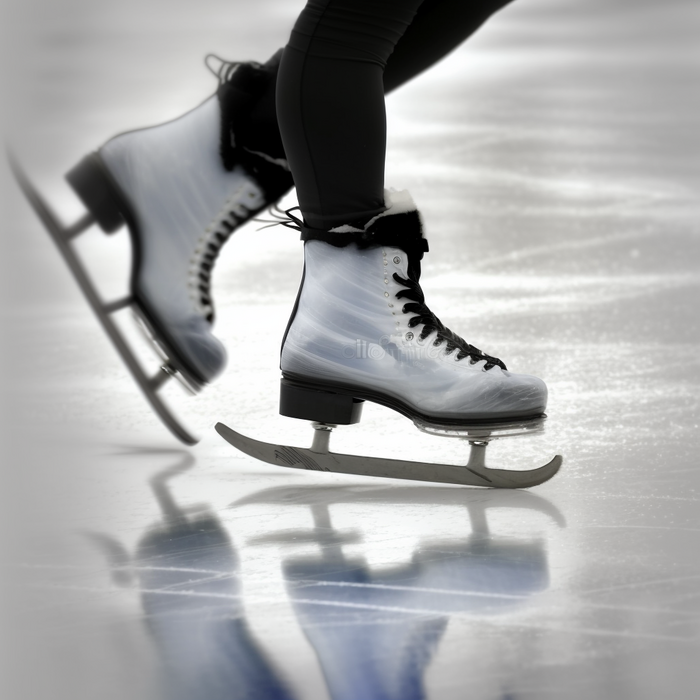}}\hfill
    \subfloat[a person is skating on an \underline{ice rink}]{\includegraphics[width=0.3\columnwidth,height=2.5cm]{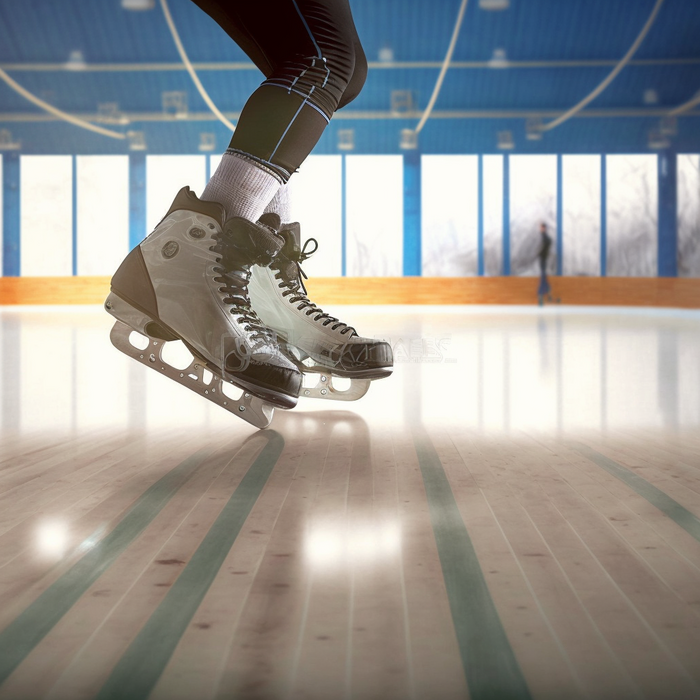}}
    \caption*{(1) Wrong caption only for the \emph{weird} image (caption \textit{c}).\\The flooring is made of wooden parquet, and not an ice rink.}
    \label{fig:subfigures1}
    \subfloat[an old abandoned house in the middle of a field with a lightning bolt]{\includegraphics[width=0.3\columnwidth,height=2.5cm]{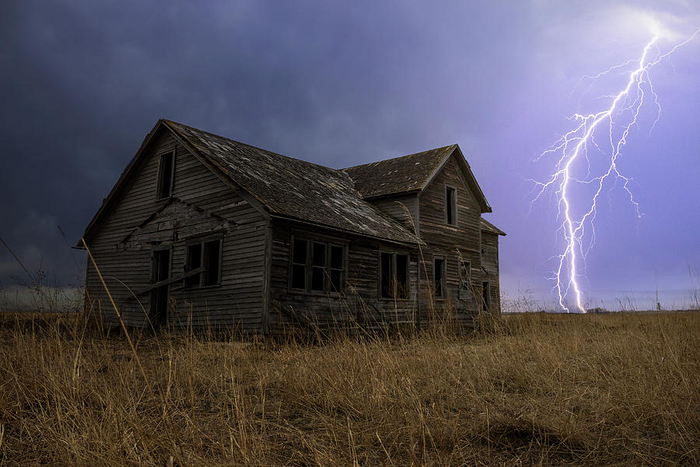}}\hfill
    \subfloat[a house in the middle of a field]{\includegraphics[width=0.3\columnwidth,height=2.5cm]{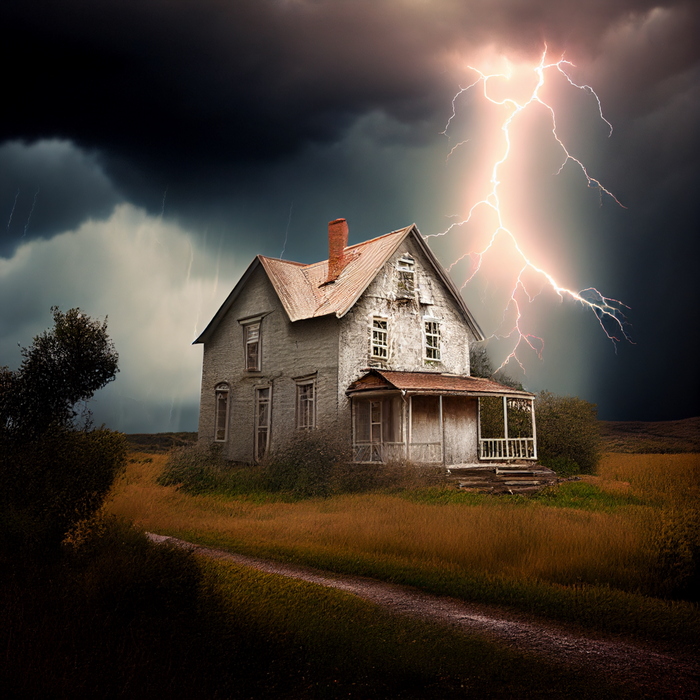}}\hfill
    \subfloat[an old house in the desert with a lightning bolt]{\includegraphics[width=0.3\columnwidth,height=2.5cm]{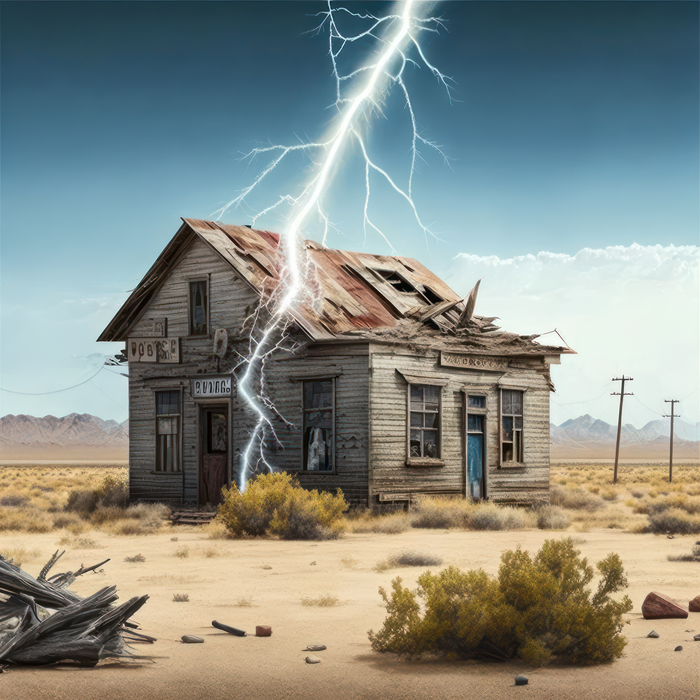}}
    \caption*{(2) Wrong captions for both synthetic images, weird and normal (captions \textit{e} and \textit{f}). The middle one misses the lightning, the right one misses the clear sky.}
    \label{fig:subfigures2}
    
    \subfloat[\underline{a man} in a top hat playing an electric guitar]{\includegraphics[width=0.3\columnwidth,height=2.5cm]{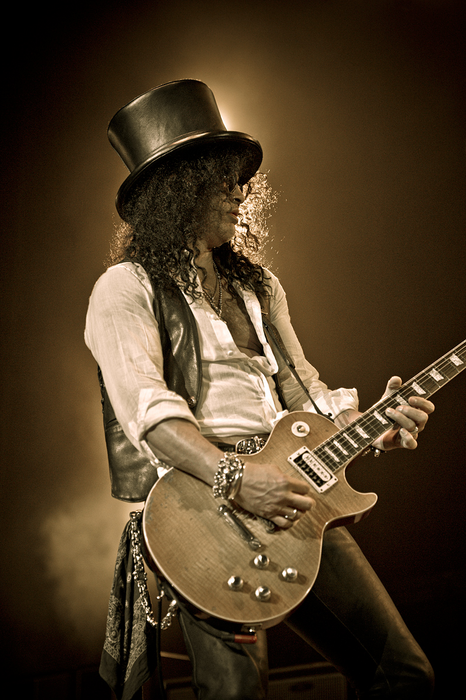}}\hfill
    \subfloat[slash plays a custom electric guitar in smoke]{\includegraphics[width=0.3\columnwidth,height=2.5cm]{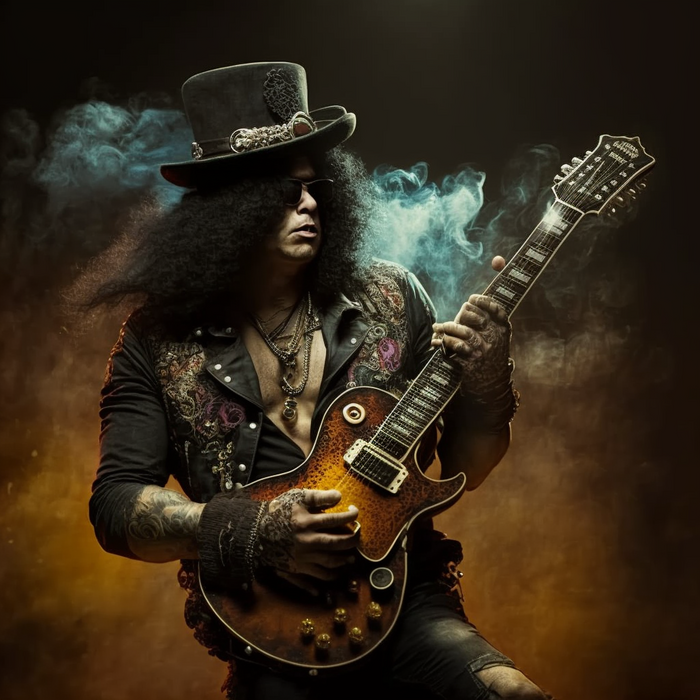}}\hfill
    \subfloat[slash playing a saxophone in a band]{\includegraphics[width=0.3\columnwidth,height=2.5cm]{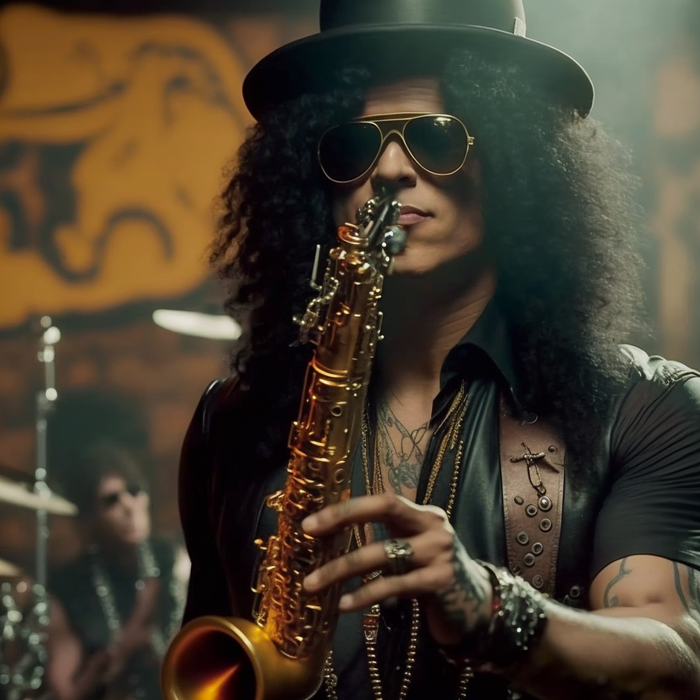}}
    \caption*{(3) Wrong caption only for the natural image (caption \textit{g}). The left caption misses the famous guitarist name (\emph{Slash})}
    \label{fig:subfigures3}
    \caption{Examples of caption errors by the BLIP2 model. The images from left to right are the \emph{weird} (synthetic) images, \emph{normal} (synthetic, without weirdness) and \emph{natural}.}
    \label{fig:analysis}
\end{figure}
\begin{table}[!t]
\centering
\begin{tabular}{@{}cccc@{}}
\toprule
Weird & Normal & Natural & \% Proportion \\ \midrule
V       & V      & V       & 45            \\ 
X       & V      & V       & 40            \\
X       & X      & X       & \phantom{0}6             \\
X       & V      & X       & \phantom{0}3             \\
V       & X      & V       & \phantom{0}2             \\
V       & V      & X       & \phantom{0}2             \\
X       & X      & V       & \phantom{0}2             \\
V       & X      & X       & \phantom{0}0             \\

\bottomrule
\end{tabular}
\caption{Analysis of caption errors, human rate of correct caption: 40\% of the errors are ``commonsense errors'' where the incorrect caption is for the ``weird'' image only. Only 3\% are ``naturalness errors'' where the natural image caption is better than the synthetic image captions. 5\% of the cases had synthetic image captions better than natural ones. }
\label{tab:analysis_truth_table}
\end{table}

\begin{figure*}[htp]
    \centering
\includegraphics[width=\textwidth]{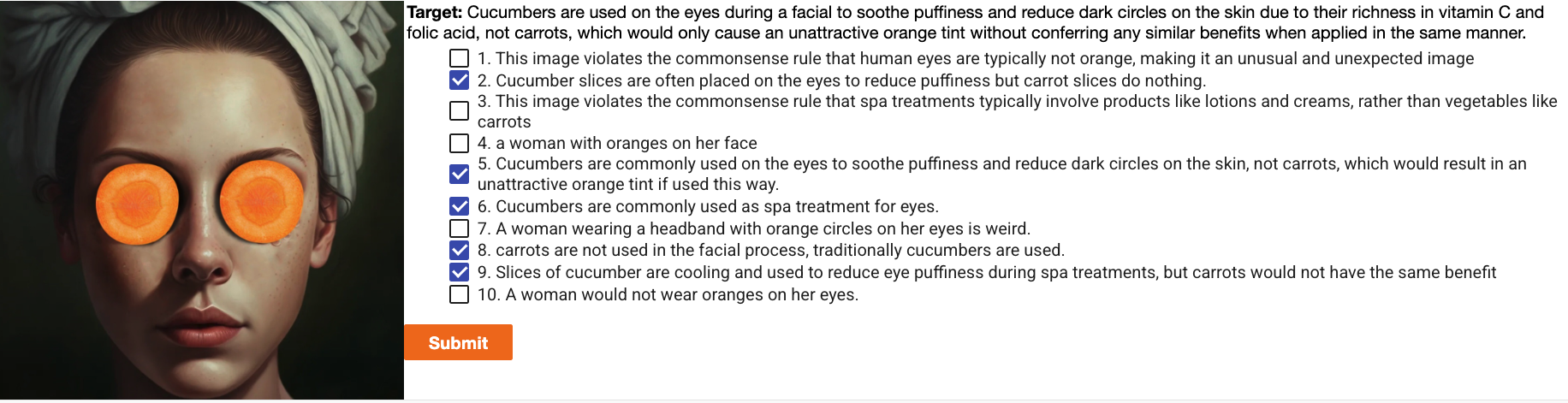}\\
    \caption{Amazon Mechanical Turk user interface for the task of explanation selection. The annotators receive an image and number of explanations, both human curated and models predictions, and need to mark the correct explanations.}
    \label{fig:explanation_selection}
\end{figure*}

In this section, we analyze if the challenges in \datasetname come from the syntheticity or weirdness of images and evaluate the performance on different commonsense categories.

\subsection{Main Challenge: Weirdness, not Synthesis}
To discern whether the difficulties models face in \datasetname{} arise from the images being ``weird'' or synthetic, we collect a set of ``normal'' and ``natural'' images. The ``normal'' images are created by replacing the unconventional element with a conventional element, resulting in minimal changes between the pairs of $(normal, weird)$ images, following the idea of contrast sets~\cite{gardner2020evaluating,bitton2021automatic}. To obtain the ``natural'' images, we search for similar non-synthetic images using Clip Retrieval~\cite{beaumont-2022-clip-retrieval}, which finds close images through CLIP \cite{radford2021learning} embedding similarity. High quality images with an ``aesthetic score'' above 7 are chosen, and the top similar images are selected. \cref{fig:analysis} shows examples of the collected images, including an image of ice skates in a ``natural'' photograph, and two synthetic images, one on a conventional element (an ice rink) and the other on an unconventional surface (a basketball parquet).

Next, we use  BLIP2 to generate captions for 300 images in three categories: natural, normal, and weird. Human annotators evaluate the accuracy of the captions for each image category, and the results are presented in \cref{tab:analysis_truth_table}. The accuracy of image captioning is high for the natural and normal categories (89\%), but low for the weird category (49\%). We find BLIP2 generates correct captions for all three categories for 45\% of the image triplets. For 40\% of the triplets, incorrect captions are generated only for the weird images (\cref{fig:analysis} (a)). In 5\% of the cases, synthetic images have better captions than natural ones, while only 3\% of cases have errors related to naturalness, with incorrect captions for both synthetic images (\cref{fig:analysis} (b)). These results suggest that the BLIP2 model can generate captions for synthetic images as well as natural images with high accuracy, but the primary challenge lies in commonsense reasoning, underscoring the need for improvement in state-of-the-art models. 
The full experiment results are available in the project website.

\subsection{Performance by Commonsense Categories} In \cref{fig:clusters}, we included the performance of the top two models, demonstrating that \datasetname provides insights into their strengths and weaknesses. Specifically, we observe that the Predicted Caption $\rightarrow$ GPT3 pipeline approach outperforms the Supervised BLIP2-XXL end-to-end model in 46\% of the categories, such as in cases of Incorrect usage (e.g., A bowl of ice cream is inside the microwave), and performs worse in 23\% of the categories, such as in Biological rules (e.g., A mouse hatches from an egg), and similarly in 31\% of the categories. Notably, both models perform poorly in identifying temporal discrepancy (e.g., women in ornate Renaissance clothing take a selfie with a smartphone) and art knowledge (e.g., The Girl with a Pearl Earring wears a golden hoop earring), while performing well in identifying an Unsuitable Environment (e.g., A snowman sits on the beach on a sunny day).

\section{Conclusions}
We introduced \datasetname, a novel dataset of synthetic images challenging AI models to reason about commonsense and compositionality. Using text-to-image models, we generated difficult or impossible to obtain images and annotated them with explanations, captions, underspecified captions, and visual question answering pairs. We proposed a benchmark of four challenging tasks and evaluated state-of-the-art models, which struggled, especially in the new task of explanation generation, where a significant gap between human and model performance remains. Our dataset and benchmark tasks are a valuable resource for advancing research in these areas. 
We provide an evaluation code and a leaderboard for methodical tracking and replication of results across our four benchmark tasks.

\section{Limitations}
We took measures to filter out potentially harmful or offensive images and texts in \datasetname (\cref{sec:toxic}), but it is still possible that some individuals may find certain content objectionable. Any harmful cases can be reported in the project website and removed from the dataset. 

While \datasetname has fewer images than other benchmarks, we intentionally created unique and challenging images to provide diverse commonsense challenges. The smaller size allowed for efficient manual annotation and evaluation, ensuring data quality and reliability. We plan to expand the dataset in the future to enhance its usefulness.

We have made significant efforts to develop reliable and advanced models for this task, our focus is not on achieving the ultimate upper bound on model performance, but on providing a challenging resource for commonsense and compositionality using image generation models.



{\small
\bibliographystyle{ieee_fullname}
\bibliography{egbib}
}

\appendix
\section{Appendix}
\label{app:appendix}

Magic icon in the title created by Freepik - Flaticon \url{https://www.flaticon.com/free-icons/magic}.

The following sections are relevant for \textbf{reproducibility} and providing more details and examples about the models evaluation, training and data collection. 

\subsection{Model versions}
\label{sec:reproduce}
We take the official implementations for CLIP: \url{https://github.com/openai/CLIP}, OFA: \url{https://github.com/OFA-Sys/OFA}, CoCa: \url{colab.research.google.com/github/mlfoundations/open_clip/blob/master/docs/Interacting_with_open_coca.ipynb}, BLIP: \url{https://colab.research.google.com/github/salesforce/BLIP/blob/main/demo.ipynb}, BLIP2: \url{https://github.com/salesforce/LAVIS/tree/main/projects/blip2}. We take the versions specified in the paper without changing the default hyper-parameters.

\subsection{Supervised Data Details}
\label{sec:sec_with_supervised_data_details}

\begin{table*}
\centering
\begin{tabular}{lp{6cm}p{6cm}}
\toprule
Task & Input ($x_{text}$) & Target ($y_{text}$) \\
\midrule
   VQA & Question: Through what is the man drinking tea looking at the Earth? & a porthole \\[1cm]
   Captioning & Describe the image. & A snow plow is plowing sand in the desert. \\[1cm]
   Matching & Which is better? A: Boys are being rained on. B: A group of children are wearing raincoats in a classroom. & B \\[1cm]
   Weird Id & Is this normal or weird? & weird \\[1cm]
   Weird Explain (crowd) & Why is it weird? & Walking in the road is dangerous, especially for a child who should be on a sidewalk instead.\\[1cm]
   Weird Explain (designer) & Why is it weird, in detail? &  For an indoor fire to be safe, it has to be adequately ventilated and contained within a fireproof environment like a fireplace or a modern stove, which is why you don't see a campfire indoors because the fire would quickly spread and destroy everything and the carbon monoxide would suffocate any living creatures. \\[1cm]
\bottomrule
\end{tabular}

\caption{Illustrations of sequence-to-sequence formatting of each task (images omitted, but are also included as inputs). Our supervised models are trained on a concatenation of all training data for all tasks.}
\label{tab:example_instances}
\end{table*}

We format all supervised tasks in a sequence-to-sequence fashion. All training examples have the format $\langle x_{image}, x_{text}, y_{text} \rangle$. Where BLIP-2 Flan-T5 is trained to maximize the probability of the textual target given the image and prompt inputs, i.e., $P(y_{text} | x_{image}, x_{text})$. For each cross-validation split, we train a single multitask model for all tasks by setting the prompt in a strategic fashion. Example input/output targets are given in~\cref{tab:example_instances}. We provide the training splits for completeness, although we recommend using \datasetname primarily as a test set.

The models were trained using 8xA6000 GPUs. A single training run for 15 epochs takes around 2 hours. To train all 5 splits, over two learning rates and three models, it requires approximately 72 hours of compute time on a machine with 8xA6000 GPUs, which is equivalent to about 576 GPU hours.

\subsection{Image Generation Designers Guidelines}
\label{sec:image_generation_guidelines}
The task is to create an image that depicts something ``weird'' that will be intuitive for humans to understand and challenging to artificial intelligence models. The images should be relatively realistic, with one weird thing that requires the use of logic and general knowledge. The goal is to compare the explanations provided by the models and people to determine whether the models struggle more than humans.

To ensure the task is challenging for AI models, the criteria for ``weird'' should be conceptual and not directly related to the rest of the picture. For example, there should be no other illogical things in the picture, such as distorted objects or more than ten fingers.

To create prompts, participants should replace $X_1$ with some $X_2$ in situations where $X$ and $Y$ appear together in a normal way in the real world. The similarity between $X_1$ and $X_2$ should mislead the models when they ask ``What's weird in the picture?'' The prompts should include cultural, general knowledge, times, and behavioral elements that make the picture illogical, but not directly related to what is happening in the picture.

Participants will receive a link to their own shared directory where they can upload their high-quality images and prompts that follow the naming conventions provided by Midjourney.\footnote{www.midjourney.com} The prompts that create the images should be recorded, and formats with seed should be used to allow the images to be restored later.

\begin{figure}[!htb]
    \centering
    \includegraphics[width=\columnwidth]{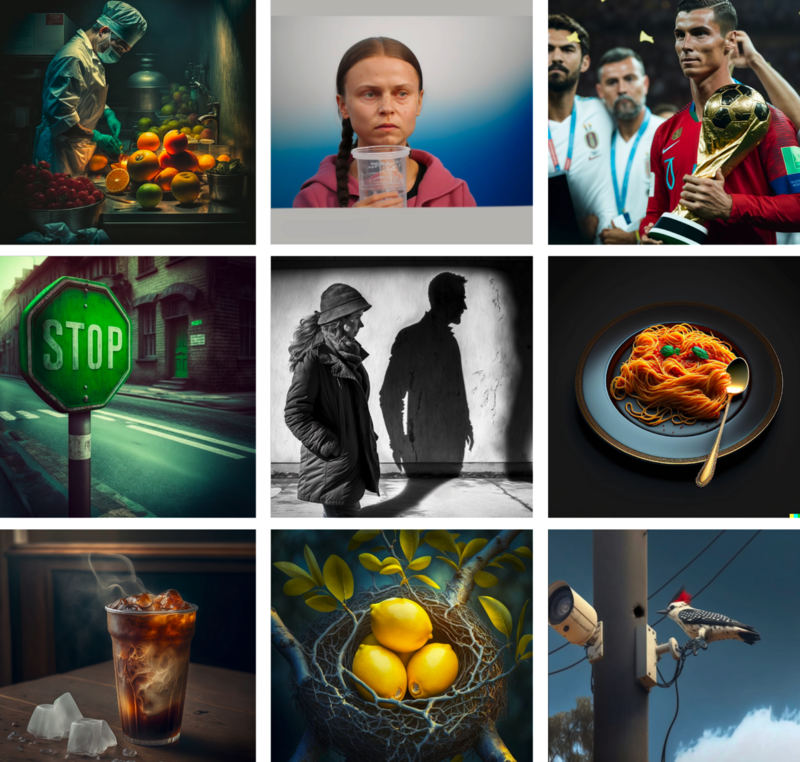}\\
    \caption{Designer Guidelines Examples.}
    \label{fig:guideline_grid}
\end{figure}

Example of prompts for the images in \cref{fig:guideline_grid}: 
\begin{enumerate}
\item Ronaldo ($X_2$) with the World Cup trophy ($Y$) instead of Messi ($X_1$) with the World Cup trophy ($Y$)
\item Greta Thunberg ($Y$) holds a disposable cup ($X_2$) instead of a reusable cup ($X_1$)
\item A surgeon ($Y$) in the kitchen with fruit ($X_2$) instead of an operating room ($X_1$)
\item Spaghetti plate ($Y$) with spoon ($X_2$) instead of fork ($X_1$)
\item Image of a girl ($Y$) with a shadow of a man ($X_2$) instead of a girl ($Y$) with her own shadow ($X_1$)
\item Green stop sign ($X_2$) on the street ($Y$) instead of a red stop sign ($X_1$)
\item Woodpecker ($Y$) makes a hole in a metal electric pole ($X_2$) instead of a tree ($X_1$)
\item Lemons ($X_2$) in nest ($Y$) instead of bird eggs ($X_1$)
\item Cup of cold coffee ($X_2$) with steam ($Y$) instead of a cup of hot coffee ($X_1$) with steam
\end{enumerate}

\subsection{Commonsense Categories}
\label{sec:commonsense_categories}
\begin{figure*}[!tb]
    \centering
    \includegraphics[width=0.8\textwidth
    ]
    {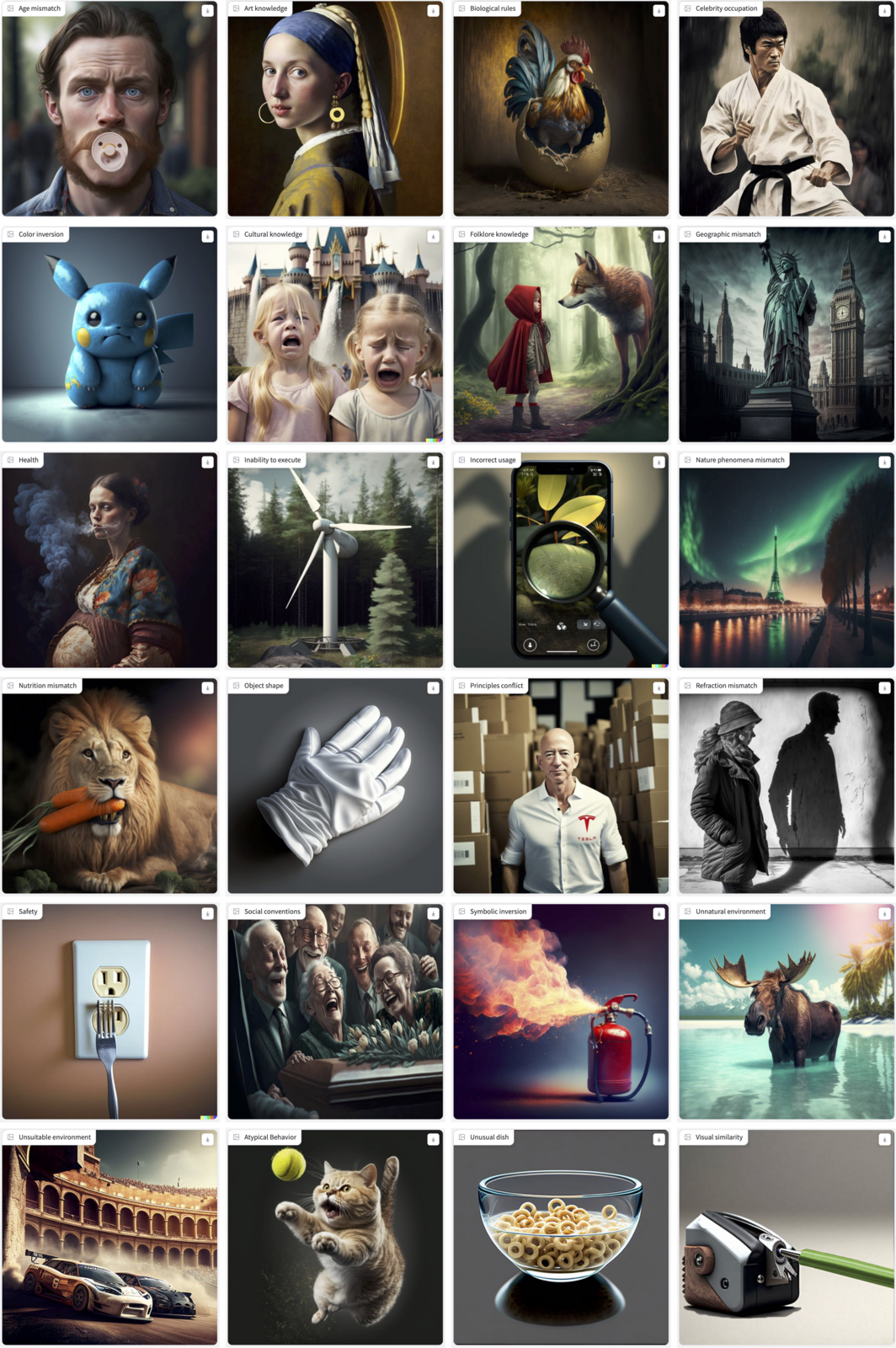}
    \caption{The WHOOPS! images span across various categories and are intended to test AI models in multiple areas of common sense.}
    \label{fig:fig_commonsense_categories}
\end{figure*}

WHOOPS! contains 26 different commonsense categories. ~\cref{fig:fig_commonsense_categories} presents examples for 24 of them, the rest can be found in the paper. In ~\cref{fig:fig1} the image of Albert Einstein is categorized as Temporal discrepancy and the candle as Physics rules. 
The concept of inability to execute refers to a scenario where an object is unable to fulfill its intended purpose due to a change or situation depicted in the image. For instance, in the image shown in~\cref{fig:fig_commonsense_categories}, the presence of trees in the forest blocks the wind from reaching the wind turbine, resulting in its inability to generate electricity.
The Unnatural Environment category pertains to instances where objects, particularly animals, are depicted in settings that are not their natural habitats, such as a moose found on a tropical beach. On the other hand, the Unsuitable Environment category refers to situations where the object is placed in a location that is not suitable for fulfilling its intended function, as seen in the example of car racing in the Colosseum.

\subsection{Human Annotation}
\label{app:human_annotation}
\begin{figure}[!htb]
    \centering
    \includegraphics[width=\columnwidth]{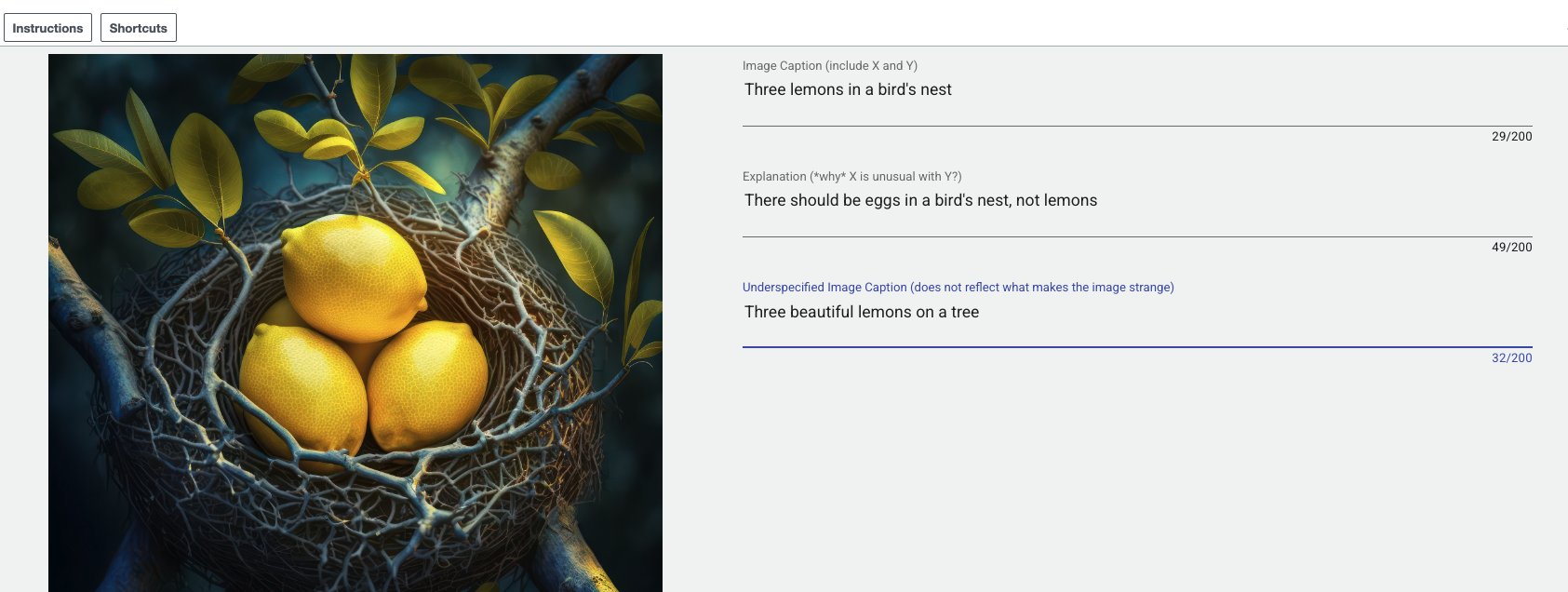}\\
    \caption{Amazon Mechanical Turk Annotators user interface. The annotators receive an image and provide three types of annotations.}
    \label{fig:mturk_ui_examples}
\end{figure}

\begin{figure}[!htb]
    \centering
    \includegraphics[width=\columnwidth]{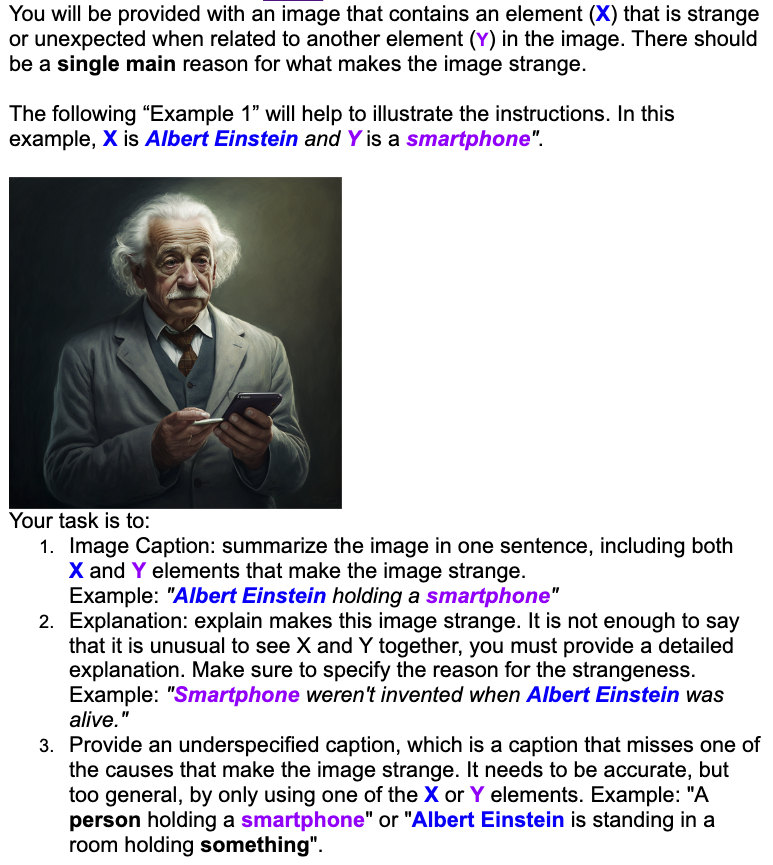}\\
    \caption{Amazon Mechanical Turk Annotators Instructions for Dataset Annotation.}
    \label{fig:fig_annotators_instructions}
\end{figure}

\cref{fig:mturk_ui_examples} shows an example of the Mechanical Turk user-interface. \cref{fig:fig_annotators_instructions} shows the instructions given to the annotators. 

The basic requirements for our annotation task is percentage of approved assignments above 98\%, more than 5,000 approved HITs, the location from the US, UK, Australia or New Zealand. We selected 5 examples from our dataset as qualification test and screen the annotators results.

\begin{figure}[!htb]
    \centering
    \includegraphics[width=\columnwidth]{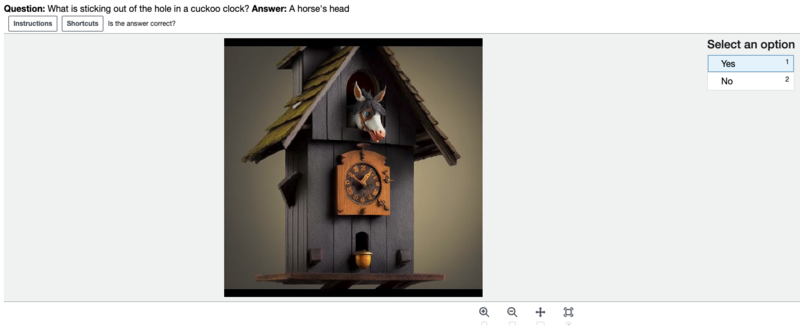}\\
    \caption{Amazon Mechanical Turk interface for the task of VQA verification. The annotators receive an (image, question, answer) triplet, and need to determine whether the answer is correct or not.}
    \label{fig:vqa_verification}
\end{figure}

\cref{fig:vqa_verification} shows an example from the VQA verification part, where the annotators are asked to determine whether a visual question answering instance generated by an automatic process is correct.

\cref{fig:explanation_selection} shows an example from the explanation selection part, where the annotators are asked to select correct explanations for the task. The options are both human selections and model predictions, and by aggregate the raters selections, we can extract human metric performance for both the human annotators who solved the task, and both for the different models. The explanation selection process is critical for obtaining accurate evaluations of model performance. The annotators are presented with a set of options that include both human and model-generated explanations. The selected explanations are then aggregated to derive a human metric for both the annotators and the models. A good explanation should identify why X and Y are unusual together due to some reason Z. For example, ``Thorns are sharp and will cut the brides arms'' correctly identifies the reason why thorns and a bride are unusual together. However, if the explanation fails to identify reason Z, it is not acceptable, such as ``An old man cannot skateboard.'' General statements or Wikipedia snippets that fail to explain what makes the image strange are also inadequate, e.g., ``Brides usually hold a bouquet of flowers''. Moreover, explanations that contain incorrect information are not considered correct. Finally, if an explanation requires verification with a search engine, it should be excluded from the selection process. Overall, the explanation selection process ensures that the chosen explanations accurately capture the reason for an image's weirdness and contribute to the accurate evaluation of model performance.

\end{document}